\documentclass[alpha-refs]{wiley-article}

\usepackage{siunitx}
\usepackage{soul,xcolor} %strikethrough with color options
\usepackage{ulem}
\papertype{Review Article}

\title{Practical and Ethical Challenges of Large Language Models in Education: A Systematic Scoping Review}

\author[1]{Lixiang Yan}
\author[1]{Lele Sha} 
\author[1]{Linxuan Zhao}
\author[1]{Yuheng Li}
\author[1]{Roberto Martinez-Maldonado} 
\author[1]{Guanliang Chen} 
\author[1]{Xinyu Li}
\author[1]{Yueqiao Jin}
\author[1]{Dragan Gašević}

\affil[1]{Centre for Learning Analytics at Monash, Faculty of Information Technology, Monash University, Clayton, Victoria, Australia}

\corraddress{Lixiang Yan, Centre for Learning Analytics at Monash, Faculty of Information Technology, Monash University, 20 Exhibition Walk, Clayton, VIC 3800, Australia}
\corremail{jimmie.yan@monash.edu}

\fundinginfo{This research was at least in part funded by the Australian Research Council (DP210100060) and Jacobs Foundation (Research Fellowship).}

\runningauthor{Yan et al.}

\begin{document}
\setstcolor{red} % set the color for strikethrough
\begin{frontmatter}
\maketitle

\begin{abstract}

Educational technology innovations leveraging large language models (LLMs) have shown the potential to automate the laborious process of generating and analysing textual content. While various innovations have been developed to automate a range of educational tasks (e.g., question generation, feedback provision, and essay grading), there are concerns regarding the practicality and ethicality of these innovations. Such concerns may hinder future research and the adoption of LLMs-based innovations in authentic educational contexts. To address this, we conducted a systematic scoping review of 118 peer-reviewed papers published since 2017 to pinpoint the current state of research on using LLMs to automate and support educational tasks. The findings revealed 53 use cases for LLMs in automating education tasks, categorised into nine main categories: profiling/labelling, detection, grading, teaching support, prediction, knowledge representation, feedback, content generation, and recommendation. Additionally, we also identified several practical and ethical challenges, including low technological readiness, lack of replicability and transparency, and insufficient privacy and beneficence considerations. The findings were summarised into three recommendations for future studies, including updating existing innovations with state-of-the-art models (e.g., GPT-3/4), embracing the initiative of open-sourcing models/systems, and adopting a human-centred approach throughout the developmental process. As the intersection of AI and education is continuously evolving, the findings of this study can serve as an essential reference point for researchers, allowing them to leverage the strengths, learn from the limitations, and uncover potential research opportunities enabled by ChatGPT and other generative AI models.

% Please include a maximum of seven keywords
\end{abstract}
\end{frontmatter}

% \section*{Author Bio}

% Lixiang Yan is a PhD candidate in the Faculty of Information Technology at Monash University. His research interests include artificial intelligence, multimodal learning analytics, collaborative learning, and applied machine learning. Roberto Martinez-Maldonado is a senior lecturer of learning analytics and human-computer interaction in the Faculty of Information Technology at Monash University. His current research is in the area of learning and teamwork analytics, in which he utilises his expertise in human-computer interaction, collaborative learning and artificial intelligence. Linxuan Zhao is a PhD candidate in the Faculty of Information Technology at Monash University. His research interests include audio analytics, multimodal learning analytics, and collaborative learning. Xinyu Li is a research fellow in the Faculty of Information Technology at Monash University. His main research areas include self-regulated learning analytics, the application of computer vision in collaborative learning and multimodal learning analytics. Yueqiao Jin is a research assistant at the Centre for Learning Analytics at Monash. Her research interests include human-centred learning analytics and artificial intelligence. Dragan Gašević is a distinguished professor of learning analytics in the Faculty of Information Technology and director of the Centre for Learning Analytics at Monash University. His research interests are learning analytics, educational data mining, self-regulated learning, and collaborative learning. 

\section*{Practitioner notes}
What is currently known about this topic
\begin{itemize}
    \item Generating and analysing text-based content are time-consuming and laborious tasks.    
    \item Large language models are capable of efficiently analysing an unprecedented amount of textual content and completing complex natural language processing and generation tasks.
    \item Large language models have been increasingly used to develop educational technologies that aim to automate the generation and analysis of textual content, such as automated question generation and essay scoring.
\end{itemize}
What this paper adds
\begin{itemize}
    \item A comprehensive list of 53 different educational tasks that could potentially benefit from LLMs-based innovations through automation.
    \item A structured assessment of the practicality and ethicality of existing LLMs-based innovations from seven important aspects using established frameworks. 
    \item Three recommendations that could potentially support future studies to develop LLMs-based innovations that are practical and ethical to implement in authentic educational contexts.
\end{itemize}
Implications for practitioners
\begin{itemize}
    \item Updating existing innovations with state-of-the-art models may further reduce the amount of manual effort required for adapting existing models to different educational tasks.
    \item The reporting standards of empirical research that aims to develop educational technologies using large language models need to be improved.
    \item Adopting a human-centred approach throughout the developmental process could contribute to resolving the practical and ethical challenges of large language models in education. 
\end{itemize}

\section{Introduction}

Advancements in generative artificial intelligence (AI) and large language models (LLMs) have fueled the development of many educational technology innovations that aim to automate the often time-consuming and laborious tasks of generating and analysing textual content (e.g., generating open-ended questions and analysing student feedback survey) \citep{kasneci2023chatgpt,wollny2021we,leiker2023prototyping}. LLMs are generative artificial intelligence models that have been trained on an extensive amount of text data, capable of generating human-like text content based on natural language inputs. Specifically, these LLMs, such as Bidirectional Encoder Representations from Transformers (BERT) \citep{devlin2018bert} and Generative Pre-trained Transformer (GPT) \citep{brown2020language}, utilise deep learning and self-attention mechanisms \citep{vaswani2017attention} to selectively attend to the different parts of input texts, depending on the focus of the current tasks, allowing the model to learn complex patterns and relationships among textual contents, such as their semantic, contextual, and syntactic relationships \citep{min2021recent,liu2023context}. As several LLMs (e.g., GPT-3 and Codex) have been pre-trained on massive amounts of data across multiple disciplines, they are capable of completing natural language processing tasks with little (few-shot learning) or no additional training (zero-shot learning) \citep{brown2020language,wu2023matching}. This could lower the technological barriers to LLMs-based innovations as researchers and practitioners can develop new educational technologies by fine-tuning LLMs on specific educational tasks without starting from scratch \citep{caines2023application,sridhar2023harnessing}. The recent release of ChatGPT, an LLMs-based generative AI chatbot that requires only natural language prompts without additional model training or fine-tuning \citep{openai2022chatgpt}, has further lowered the barrier for individuals without technological background to leverage the generative powers of LLMs.

Although educational research that leverages LLMs to develop technological innovations for automating educational tasks is yet to achieve its full potential (i.e., most works have focused on improving model performances \citep{kurdi2020systematic,ramesh2022automated}), a growing body of literature hints at how different stakeholders could potentially benefit from such innovations. Specifically, these innovations could potentially play a vital role in addressing teachers' high levels of stress and burnout by reducing their heavy workloads by automating punctual, time-consuming tasks \citep{carroll2022teacher} such as question generation \citep{kurdi2020systematic,bulut9automatic,oleny2023generating}, feedback provision \citep{cavalcanti2021automatic,nye2023generative}, scoring essays \citep{ramesh2022automated} and short answers \citep{zeng2023Curriculum}. These innovations could also potentially benefit both students and institutions by improving the efficiency of often tedious administrative processes such as learning resource recommendation, course recommendation and student feedback evaluation, potentially  \citep{zawacki2019systematic,wollny2021we,sridhar2023harnessing}. 

Despite the growing empirical evidence of LLMs' potential in automating a wide range of educational tasks, none of the existing work has systematically reviewed the practical and ethical challenges of these LLMs-based innovations. Understanding these challenges is essential for developing responsible technologies as LLMs-based innovations (e.g., ChatGPT) could contain human-like biases based on the existing ethical and moral norms of society, such as inheriting biased and toxic knowledge (e.g., gender and racial biases) when trained on unfiltered internet text data \citep{schramowski2022large}. Prior systematic reviews have focused on investigating these issues related to one specific application scenario of LLMs-based innovations (e.g.,  question generation, essay scoring, chatbots, or automated feedback) \citep{kurdi2020systematic,cavalcanti2021automatic,wollny2021we,ramesh2022automated}. The practical and ethical challenges of LLMs in automating different types of educational tasks remain unclear. Understanding these challenges is essential for translating research findings into educational technologies that stakeholders (e.g., students, teachers, and institutions) can use in authentic teaching and learning practices \citep{adams2021artificial}.

The current study is the first systematic scoping review that aimed to address this gap by reviewing the \textit{current state of research} on using LLMs to automate educational tasks and identify the \textit{practical} and \textit{ethical} challenges of adopting these LLMs-based innovations in authentic educational contexts. A total of 118 peer-reviewed publications from four prominent databases were included in this review following the Preferred Reporting Items for Systematic Reviews and Meta-Analyses (PRISMA) \citep{page2021prisma} protocol. An inductive thematic analysis was conducted to extract details regarding the different types of educational tasks, stakeholders, LLMs, and machine learning tasks investigated in prior literature. The practicality of LLMs-based innovations was assessed through the lens of technological readiness, model performance, and model replicability. Lastly, the ethicality of these innovations was assessed by investigating system transparency, privacy, equality, and beneficence. 

The contribution of this paper to the educational technology community is threefold: 1) we systematically summarise a comprehensive list of 53 different educational tasks that could potentially benefit from LLMs-based innovations through automation, 2) we present a structured assessment of the practicality and ethicality of existing LLMs-based innovations based on seven important aspects using established frameworks (e.g., the transparency index \citep{chaudhry2022transparency}), and 3) we propose three recommendations that could potentially support future studies to develop LLMs-based innovations to be practically and ethically implement in authentic educational contexts. As the intersection of LLMs and education is continuously evolving, the findings of this systematic scoping review can serve as an essential reference point for researchers, allowing them to leverage the strengths, learn from the limitations, and uncover potential opportunities for novel LLMs in supporting educational research and practice. Specifically, emerging works should carefully consider the practical and ethical challenges identified in this study while exploring the research opportunities enabled by ChatGPT and other generative AI models.

\section{Background}
\label{Definitions}

In this section, we first establish the definitions for the key terminologies, specifically the definitions of practicality and ethicality in the context of educational technology. We then provided an overview of prior systematic reviews on LLMs in education. Then, we present the research questions based on the gaps identified in the existing literature.

\subsection{Practicality}

Several theoretical frameworks have been proposed regarding the practicality of integrating technological innovations in educational settings. For example, Ertmer's (\citeyear{ertmer1999addressing}) first- and second-order barriers to change focused on the external conditions of the educational system (e.g., infrastructure readiness) and teachers' internal states (e.g., personal beliefs). \citet{becker2000findings} further suggested that for technological innovations to have actual benefits in supporting pedagogical practices, these innovations should be convenient to access, support constructivist pedagogical beliefs, be adaptable to changes in the curriculum, and be compatible to teachers' level of knowledge and skills. These factors were also presented in an earlier framework of the practicality index \citep{doyle1977practicality}, which summarised three critical components for integrating educational technologies, including the degree of adoption feasibility, the cost and benefit ratio, and the alignment with existing practices and beliefs. Based on these prior theoretical frameworks and considering the recentness of LLMs-based innovations (which only emerged in the past five years), the practical challenges of LLMs-based innovations in automating educational tasks can be assessed from three primary perspectives. First, evaluating the technological readiness of these innovations is essential for determining whether there is empirical evidence to support successful integration and operation in authentic educational contexts. Second, assessing the model performance could contribute valuable insights into the cost and benefits of adopting these innovations, such as comparing the benefits of automation with the costs of inaccurate predictions. Finally, understanding whether these innovations are methodologically replicable could be important for future studies to investigate their alignment with different educational contexts and stakeholders. We elaborated on the evaluation items for each challenge in Section \ref{sec:analysis}. 

\subsection{Ethicality}

Ethical AI is a prevalent topic of discussion in multiple communities, such as learning analytics, AI in education, educational data mining, and educational technology communities \citep{adams2021artificial,pardo2014ethical}. There are ongoing debates regarding AI ethics in education with a mixture of focuses on algorithmic and human ethics among educational data mining and AI in education communities \citep{holmes2022ethics}. As such debates continue, it is difficult to identify an established definition of ethical AI from these fields. Whereas, ethicality has already been thoroughly investigated and addressed in a closed field to AI in education, namely, the field of learning analytics \citep{pardo2014ethical,selwyn2019s}. Drawing on the established definition of ethicality from the field of learning analytics \citep{pardo2014ethical}, the ethicality of LLMs-based innovations can thus be defined as the systematisation of appropriate and inappropriate functionalities and outcomes of these innovations, as determined by all stakeholders (e.g., students, teachers, parents, and institutions). For example, \citet{khosravi2022explainable} explained that the ethicality of AI-powered educational technology systems needs to involve the consideration of accountability, explainability, fairness, interpretability, and safety of these systems.
%Recent studies on the ethicality of AI-powered educational technology systems often focused on the accountability, explainability, fairness, interpretability, and safety of these systems \citep{khosravi2022explainable}. 
These different domains of ethical AI are all closely related and can be addressed by considering system transparency. Transparency is a subset of ethical AI that involves making all information, decisions, decision-making processes, and assumptions available to stakeholders, which in turn enhances their comprehension of the AI systems and related outputs \citep{chaudhry2022transparency}. Additionally, for LLMs-based innovations, \citet{weidinger2021ethical} suggested six types of ethical risks, including 1) discrimination, exclusion, and toxicity, 2) information hazards, 3) misinformation harms, 4) malicious uses, 5) human-computer interaction harms, and 6) automation, access, and environmental harms. These risks can be further aggregated into three fundamental ethical issues, such as privacy concerns regarding educational stakeholders' personal data, equality concerns regarding the accessibility of stakeholders with different backgrounds, and beneficence concerns about the potential harms and negative impacts that LLMs-based innovations may have on stakeholders \citep{ferguson2016guest}. These three fundamental ethical issues were considered in the analysis of the reviewed literature. Further details were available in Section \ref{sec:analysis}. 

\subsection{Related Work}

Prior systematic reviews have focused primarily on reviewing a specific application scenario (e.g., question generation, automated feedback, chatbots and essay scoring) of natural language processing and LLMs. For example, \citet{kurdi2020systematic} have systematically reviewed empirical studies that aimed to tackle the problem of automatic question generation in educational domains. They comprehensively summarised the different generation methods, generation tasks, and evaluation methods presented in prior literature. In particular, LLMs could potentially benefit the semantic-based approaches for generating meaningful questions that are closely related to the source contents. Likewise, \citet{cavalcanti2021automatic} have systematically reviewed different automated feedback systems regarding their impacts on improving students' learning performances and reducing teachers' workloads. Despite half of their reviewed studies showing no evidence of reducing teachers' workloads, as these automated feedback systems were mostly rule-based and required extensive manual efforts, they identified that using natural language generation techniques could further enhance such systems' generalisability and potentially reduce manual workloads. On the other hand, \citet{wollny2021we} have systematically reviewed areas of education where chatbots have already been applied. They concluded that there is still much to be done for chatbots to achieve their full potential, such as making them more adaptable to different educational contexts. A systematic review has also investigated the various automated essay scoring systems \citep{ramesh2022automated}. The findings have revealed multiple limitations of the existing systems based on traditional machine learning (e.g., regression and random forest) and deep learning algorithms (e.g., LSTM and BERT). In sum, these previous systematic reviews have identified room for improvement that can be potentially addressed using state-of-the-art LLMs (e.g., GPT-3 or Codex). However, none of the prior systematic reviews has investigated the practical and ethical issues related to LLMs-based innovations in education generally rather than particularly (e.g., limited to a specific task).

The recent hype around one of the latest LLMs-based innovations, ChatGPT, has intensified the discussion about the practical and ethical challenges related to using LLMs in education. For example, in a position paper, \citet{kasneci2023chatgpt} provided an overview of some existing LLMs research and proposed several practical opportunities and challenges of LLMs from students' and teachers' perspectives. Likewise, \citet{rudolph2023chatgpt} also provided an overview of the potential impacts, challenges, and opportunities that ChatGPT might have on future educational practices. Although these studies have not systematically reviewed the existing educational literature on LLMs, their arguments resonated with some of the pressing issues around LLMs and ethical AI, such as data privacy, bias, and risks. On the other hand, \citet{sallam2023utility} systematically reviewed the implications and limitations of ChatGPT in healthcare education and identified potential utility around personalisation and automation. However, it is worth noting that most papers reviewed in Sallam's study were either editorials, commentaries, or preprints. This lack of peer-reviewed empirical studies on ChatGPT is understandable as it has only been released since late 2022 \citep{openai2022chatgpt}. None of the existing work has systematically reviewed the peer-reviewed literature on prior LLMs-based innovations. Such investigations could provide more reliable and empirically-based evidence regarding the potential opportunities and challenges of LLMs in educational practices. Thus, the current study aimed to address this gap in the literature by conducting a systematic scoping review of prior educational research on LLMs. Specifically, the following research questions were investigated to guide this review: 

\begin{itemize}
\item \noindent\textbf{RQ1}: What is \textit{the current state of research} on using LLMs to automate educational tasks, specifically through the lens of educational tasks, stakeholders, LLMs, and machine-learning tasks\footnote[1]{Such as classification, prediction, clustering, etc.}?

\item \noindent\textbf{RQ2}: What are the \textit{practical} challenges of LLMs in automating educational tasks, specifically through the lens of technological readiness, model performance, and model replicability?

\item \noindent\textbf{RQ3}: What are the \textit{ethical} challenges of LLMs in automating educational tasks, specifically through the lens of system transparency, privacy, equality, and beneficence?

\end{itemize}

\section{Methods}

A systematic scoping review was conducted in this study as this method has been frequently used in emerging and rapidly evolving research areas to scope a body of literature and identify the key concepts, methods, evidence, and challenges \citep{munn2018systematic}. Consequently, the quality of the included studies was often not assessed as the aim is to provide a boarder picture of an emerging field.

\subsection{Review Procedures}
We followed the PRISMA \citep{page2021prisma} protocol to conduct the current systematic scoping review of LLMs. We searched four reputable bibliographic databases, including Scopus, ACM Digital Library, IEEE Xplore, and Web of Science, to find high-quality peer-reviewed publications. Additional searches were conducted through Google Scholar and Education Resources Information Center (ERIC) to identify peer-reviewed publications that have yet to be indexed by these databases, either recently published or not indexed (e.g., Journal of Educational Data Mining; prior to 2020). Our initial search query for the title, abstract, and keywords included terms such as "large language model", "pre*trained language model", "GPT-*", "BERT", "education", "student*", and "teacher*". A publication year constraint was also applied to restrict the search to studies published since 2017, specifically from 01/01/2017 to 12/31/2022, as the foundational architecture (Transformer) of LLMs was formally released in 2017 \citep{vaswani2017attention}. Only peer-reviewed publications were considered to enhance the scientific credibility of this review. The initial database search was conducted by two researchers independently. Any discrepancies between the search results were resolved through further discussion or consulting the librarian for guidance.

Two researchers independently reviewed the titles and abstracts of eligible articles based on five predetermined inclusion and exclusion criteria. First, we included studies that used large or pre-trained language models directly or built on top of such models, and excluded studies that used general machine-learning or deep-learning models with unspecified usage of LLMs. Second, we included empirical studies with detailed methodologies, such as a detailed description of the LLMs and research procedures, and excluded review, opinion, and scoping works. Third, we only included full-length peer-reviewed papers, and excluded short, workshop, and poster papers that were less than six and eight pages for double- and single-column layouts, respectively. Additionally, we included studies that used LLMs for the purpose of automating educational tasks (e.g., essay grading and question generation), and excluded studies that merely used LLMs as part of the analysis without educational implications. Finally, we only included studies that were published in English (both the abstract and the main text) and excluded studies that were published in other languages. Any conflicting decisions were resolved through further discussion between the two researchers or consulting with a third researcher to achieve a consensus.

\begin{figure}[htbp]
    \centering
      \includegraphics[width=.99\textwidth]{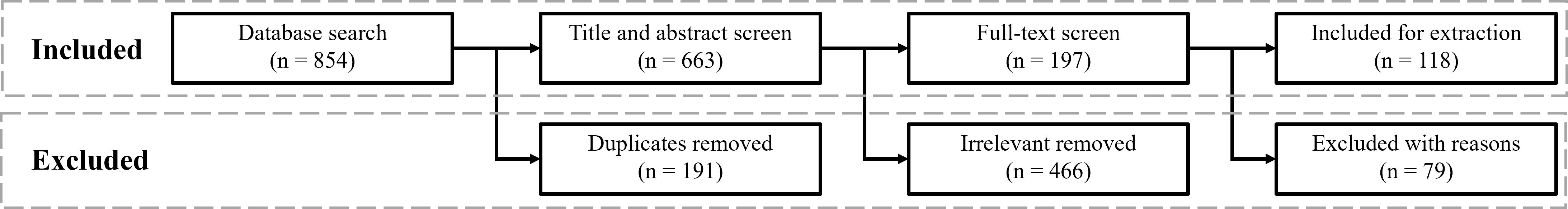}
    \caption{Systematic scoping review process following the PRISMA protocol.}
    \label{fig:prisma}
\end{figure}

The database search initially yielded 854 publications, with 191 duplicates removed, resulting in 663 publications for the title and abstract screening (see Figure \ref{fig:prisma}). After the title and abstract screening, 197 articles were included for the full-text review with an inter-rater reliability (Cohen's kappa) of 0.75, indicating substantial agreement between the reviewers during the title and abstract screening. A total of 118 articles were selected for data extraction after the full-text review with an inter-rater reliability (Cohen's kappa) of 0.73, indicating substantial agreement between the reviewers during the full-text review. Out of the initial 197 articles, 79 were excluded for various reasons, including not full paper (n=41), lack of educational automation (n=17), lack of pre-trained or LLMs (n=12), merely using pre-trained or LLMs as part of the analysis (n=3), non-English paper (n=2), and non-empirical paper (n=2). 

\subsection{Data Analysis}
\label{sec:analysis}

For the first research question (RQ1), we conducted an inductive thematic analysis to extract information regarding the current state of research on using LLMs to automate educational tasks. Specifically, we extracted four primary types of contextual information from each included paper: educational tasks, stakeholders, LLMs, and machine-learning tasks. This contextual information would provide a holistic view of the existing research and inform researchers and practitioners regarding the viable directions to explore with the state-of-the-art LLMs (e.g., GPT-3.5 and Codex). A total of seven data extraction items were developed to address the second and third research questions. These items were developed as they are directly related to the definition of practicality (RQ2: Item 1-3) and ethicality (RQ3: Item 4-7), as defined in the Background section (Section \ref{Definitions}). The following list elaborates on the final set of items along with the corresponding guiding questions. For the thematic analysis and Items, two researchers independently coded 20 random samples of the included studies. Any conflicts were resolved through further discussion or consulting a third researcher. After reaching a Cohen's kappa of more than 0.80 (indicating almost perfect agreement), each researcher coded half of the remaining 98 studies (49 studies each) and cross-checked each other's work. The database of the studies included in this review and the extracted data for each item are available in the supplementary document.
\begin{enumerate}

    \item \textbf{Technology readiness} What levels of technology readiness are the LLMs-based innovations at? We adopted the assessment tool from the Australian government, namely the Australian Department of Defence's Technology Readiness Levels (TRL) \citep{TRL2020}, which has been used to assess the maturity of educational technologies in prior SLR \citep{yan2022scalability}. There are nine different technological readiness levels: Basic Research (TRL-1), Applied Research (TRL-2), Critical Function or Proof of Concept Established (TRL-3), Lab Testing/Validation of Alpha Prototype Component/Process (TRL-4), Laboratory Testing of Integrated/Semi-Integrated System (TRL-5), Prototype System Verified (TRL-6), Integrated Pilot System Demonstrated (TRL-7), System Incorporated in Commercial Design (TRL-8), and System Proven and Ready for Full Commercial Deployment (TRL-9), further explained in the Result section.  
    
    \item \textbf{Performance}: How accurate and reliable can the LLMs-based innovations complete the designated educational tasks? For example, what are the model performance scores for classification (e.g., AUC and F1 scores), generation (e.g., BLEU score), and prediction tasks (e.g., RMSE and Pearson's correlation)? 

    \item \textbf{Replicability}: Can other researchers or practitioners replicate the LLMs-based innovations without additional support from the original authors? This item evaluates whether the paper provided sufficient details about the LLMs (e.g., open-sourced algorithms) and the dataset (e.g., open-source data).      

    \item \textbf{Transparency}: What tiers of transparency index \citep{chaudhry2022transparency} are the LLMs-based innovations at? The transparency index proposed three tiers of transparency, including transparent to AI researchers and practitioners (Tier 1), transparent to educational technology experts and enthusiasts (Tier 2), and transparent to educators and parents (Tier 3). The tier of transparency increases as educational stakeholders become fully involved in developing and evaluating the AI system. These tiers were further elaborated on in the Results section.  

    \item \textbf{Privacy}: Has the paper mentioned or considered privacy issues of their innovations? This item explores potential issues related to informed consent, transparent data collection, individuals' control over personal data, and unintended surveillance \citep{ferguson2016guest,tsai2020privacy}.
    
    \item \textbf{Equality}: Has the paper mentioned or considered equal access to their innovations? This item explores potential issues related to limited access for students from low-income backgrounds or rural areas and the linguistic limitation of the innovations, such as their capability to analyse different languages \citep{ferguson2016guest}.
    
    \item \textbf{Beneficence}: Has the paper mentioned or considered potential issues that violate the ethical principle of beneficence? Such violations may include the risks associated with labelling and profiling students, inadequate usage of machine-generated content for assessments, and algorithmic biases \citep{ferguson2016guest,zawacki2019systematic}.
    
\end{enumerate}
\section{Results}

\subsection{The Current State --- RQ1}

We identified nine different categories of educational tasks that prior studies have attempted to automate using LLMs (as shown in Table \ref{tab:educational-tasks}). Prior studies have used LLMs to automate the profiling and labelling of 17 types of education-related contents and concepts (e.g., forum posts, student sentiment, and discipline similarity), the detection of six latent constructs (e.g., confusion and urgency), the grading of five types of assessments (e.g., short answer questions and essays), the development of five types of teaching support (e.g., conversation agent and intelligent question-answering), the prediction of five types of student-orientated metrics (e.g., dropout and engagement), the construction of four types of knowledge representations (e.g., knowledge graph and entity recognition), the provision of four different forms of feedback (e.g., real-time and post-hoc feedback), the generation of four types of content (e.g., MCQs and open-ended questions), and the delivery of three types of recommendations (e.g., resource and course). Of the 118 reviewed studies, 85 studies aimed to automate educational tasks related to teachers (e.g., question grading and generation), 54 studies targeted student-related activities (e.g., feedback and resource recommendation), 20 studies focused on supporting institutional practices (e.g., course recommendations and discipline planning), and 14 studies empowered researchers with automated methods to investigate latent constructs (e.g., student confusion) and capture verbal data (e.g., speech recognition). 

We identified five categories of LLMs used in prior studies to automate educational tasks. BERT and its variations (e.g., RoBERTa, DistilBERT, multilingual BERT, LaBSE, EstBERT, and Sentence-BERT) were the most predominant model used in 109 reviewed studies. However, they often required manual effort for fine-tuning (n=90). GPT-2 and GPT-3 have been used in five and three studies, respectively. Specifically, GPT-2 and GPT-3 have performed better than BERT-based models in content generation and evaluation tasks, such as generating university math problems \citep{drori2022neural} and evaluating the quality of student-generated short answer questions \citep{moore2022assessing}. OpenAI's Codex has been used in two prior studies, specifically for code generation tasks. T5 has also been used in two prior studies for classification and generation purposes. In terms of machine-learning tasks, 74 studies used LLMs to perform classification tasks. Generation and prediction tasks were investigated in 24 and 23 prior studies, respectively. In sum, LLMs-based innovations have already been used to automate a range of educational tasks, but most of these innovations were developed on older models, such as BERT and GPT-2. Although state-of-the-art models, such as GPT-3, have been introduced for over two years \citep{brown2020language}, they have yet to be widely applied to automate educational tasks. A potential reason for this lack of adoption could be these models' commercial and close-sourced nature, increasing the financial burdens of developing and operating educational technology innovations on top of such models.

\begin{table}[htbp]
\centering
\caption{Educational Tasks in LLMs Research}
\begin{tabular}{|p{3cm}|p{9.5cm}|}
\hline
Categories &
  Educational Tasks \\
  \hline
Profiling and Labelling &
  Forum post classification, dialogue act classification, classification of learning designs, review sentiment analysis, topic modelling, pedagogical classification of MOOCs, collaborative problem-solving modelling, paraphrase quality, speech tagging, labelling educational content with knowledge components, key sentence and keyword extraction, reflective writing analysis, multimodal representational thinking, discipline similarity, concept classification, cognitive level classification, essay arguments segmentation \\
Detection &
  Semantic analyses, detecting off-task messages, confusion detection, urgency detection, conversational intent detection, teachers' behaviour detection \\
Assessment and Grading &
  Formative and summative assessment grading, short answer grading, essay grading, subjective question grading, student self-explanation \\  
Teaching Support &
  Classroom teaching, learning community support, online learning conversation agent, intelligent question-answering, teacher activity recognition \\
Prediction &
  Student performance prediction, student dropout prediction, emotional and cognitive engagement detection, growth and development indicators for college students, at-risk student identification \\
Knowledge Representation &
  Knowledge graph construction, knowledge entity recognition, knowledge tracing, cause-effect relation extraction \\  
Feedback &
  Real-time feedback, post-hoc feedback, aggregated feedback, feedback on feedback (peer-review comments) \\
Content Generation &
  MCQs generation, open-ended question generation, code generation, reply (natural language) generation \\
Recommendation &
  English reference selection and recommendation, resource recommendation, course recommendation \\
\hline
\end{tabular}
\label{tab:educational-tasks}
\end{table}

\subsection{Practical Challenges --- RQ2}

\subsubsection{Technology readiness}

According to the Technology Readiness Level scale \citep{TRL2020}, the LLMs-based innovations are still in the early development and testing stage. Over three-quarters of the LLMs studies (n=89) are in the applied research stage (TRL-2), which aims to experiment with the capability of LLMs in automating different educational tasks by developing different models and combining LLMs with other machine-learning and deep-learning techniques (e.g., RCNN \citep{shang2022representation}). Thirteen studies have established a proof of concept and demonstrated the feasibility of using LLMs-based innovations to automate certain processes of educational tasks (TRL-3). Nine studies have developed functional prototypes and conducted preliminary validation under controlled laboratory settings (TRL-4), often involving stakeholders (e.g., students and teachers) to test and evaluate the output of their innovations. Only seven studies have taken a further step and conducted validation studies in authentic learning environments, with most functional components integrated into the educational tasks (TRL-5), such as an intelligent virtual standard patient for medical students training \citep{song2022intelligent} and an intelligent chatbot for university admission \citep{nguyen2021neu}. Yet, none of the existing LLMs-based innovations has been verified through successful operations (TRL-6). Together, these findings suggest although existing LLMs-based innovations can be used to automate certain educational tasks, they have yet to show evidence regarding improvements to teaching, learning, and administrative processes in authentic educational practices.

\subsubsection{Performance}

The performance of LLMs-based innovations varies across different machine-learning and educational tasks. For classification tasks, LLMs-based innovations have shown high performance for simple educational tasks, such as modelling the topics from a list of programming assignments (best F1 = 0.95) \citep{fonseca2020automatic}, analysing the sentiment of student feedback (best F1 = 0.94) \citep{truong2020sentiment}, constructing subject knowledge graph from teaching materials (best F1 = 0.94) \citep{su2020automatic}, and classifying educational forum posts \citep{sha2022latest} (best F1 = 0.92). However, the classification performance of LLMs-based innovations decreases for other educational tasks. For example, the F1 scores for detecting student confusion in the course forum \citep{geller2021new} and students' off-task messages in game-based collaborative learning \citep{carpenter2020detecting} are around 0.77 and 0.67, respectively. Likewise, the F1 score for classifying short-answer responses varies between 0.61 to 0.82, with the lower performance on out-of-sample questions (best F1 = 0.61) \citep{condor2021automatic}. Similar performances were also observed in classifying students' argumentative essays (best F1 = 0.66) \citep{ghosh2020exploratory}. 

For prediction tasks, LLMs-based innovations have demonstrated reliable performance compared to ground truth or human raters. For example, LLMs-based innovations have achieved high scores of quadratic weighted kappa (QWK) in essay scoring, specifically for off-topic (QWK = 0.80), gibberish (QWK = 0.80), and paraphrased answers (QWK = 0.94), indicating substantial to almost perfect agreements with human raters \citep{doewes2021limitations}. Similar performances on essay scoring have been observed in several other studies (e.g., 0.80 QWK in \citep{beseiso2021novel} and 0.81 QWK in \citep{sharma2021feature}). Likewise, LLMs-based innovations' performances on automatic short-answer grading were also highly correlated with human ratings (Pearson's correlation between 0.75 to 0.82) \citep{ahmed2022application,sawatzki2022deep}.

Regarding generation tasks, LLMs-based innovations demonstrated high performance across different educational tasks. For example, LLMs-based innovations have achieved an F1 score of 0.92 for generating MCQs with single-word answers \citep{kumar2022identification}. Educational technologies developed by fine-tuning Codex also demonstrated the capability of resolving 81\% of the advanced mathematics problems \citep{drori2022neural}. Text summaries generated using BERT had no significant differences compared with student-generated summaries and can not be differentiated by graduate students \citep{merine2022risks}. Similarly, BERT-generated doctor-patient dialogues were also found to be indistinguishable from actual doctor-patient dialogues, which can be used to create virtual standard patients for medical students' diagnosis practice training \citep{song2022intelligent}. Additionally, for introductory programming courses, the state-of-the-art LLMs, Codex, could generate sensible and novel exercises for students along with an appropriate sample solution (around three out of four times) and accurate code explanation (67\% accuracy) \citep{sarsa2022automatic}. 

In sum, although the classification performance of LLMs-based innovations on complex educational tasks is far from suitable for practical adoption, LLMs-based innovations have already shown high performance on several relatively simple classification tasks that could potentially be deployed to automatically generate meaningful insights that could be useful to teachers and institutions, such as navigating through numerous student feedback and course review. Likewise, LLMs-based innovations' prediction and generation performance reveals a promising future of potentially automating the generation of educational content and the initial grading of student assessments. However, ethical issues must be considered for such implementations, which we covered in the findings for RQ3. 

\subsubsection{Replicability}

Most reviewed studies (n=107) have not disclosed sufficient details about their methodologies for other researchers and practitioners to replicate their proposed LLMs-based innovations. Among these studies, 12 studies have open-sourced the original code for developing the innovations but failed to open-source the data they used. In contrast, 20 studies have open-sourced the data they used but failed to release the actual code. Around two-thirds of the reviewed studies (n=75) have failed to release both the original code and the data they used, leaving only 11 studies publicly available for other researchers and practitioners to replicate without needing to contact the original authors. This lack of replicability could become a vital barrier to adoption, as 87 out of the 107 non-replicable studies required fine-tuning the LLMs to achieve the reported performance. This replication issue also limits others from further evaluating the generalisability of the proposed LLMs-based innovations in other datasets, constraining potential practical utilities.

\subsection{Ethical Challenges --- RQ3}

\subsubsection{Transparency}

Based on the transparency index and the three tiers of transparency \citep{chaudhry2022transparency}, most of the reviewed study reached at-most Tier 1 (n=109), which is merely considered transparent to AI researchers and practitioners. Although these studies reported details regarding their machine learning models (e.g., optimisation and hyperparameters), such information is unlikely to be interpretable and considered transparent for individuals without a strong background in machine learning. For the remaining nine studies, they reached at-most Tier 2 as they often involved some form of human-in-the-loop elements. Specifically, making the LLMs innovations available for student evaluation has been found in three studies \citep{nguyen2021neu,song2022intelligent,merine2022risks}. Such evaluations often involved students differentiating AI-generated from human-generated content \citep{song2022intelligent,merine2022risks} and assessing student satisfaction with AI-generated responses \citep{nguyen2021neu}. Likewise, two studies have involved experts in evaluating specific features of the content generated by the LLMs-based innovations, such as informativeness \citep{maheen2022automatic} and cognitive level \citep{moore2022assessing}. Surveys have been used to evaluate students' experience with LLMs-based innovations from multiple perspectives, such as the quality and difficulty of AI-generated questions \citep{drori2022neural,li2021natural} and potential learning benefits of the systems \citep{jayaraman2022effectiveness}. Finally, semi-structured interviews have been conducted to understand students' perception of the LLM system after using the system in authentic computer-supported collaborative learning activities \citep{zheng2022effects}. Although these nine studies had some elements of human-in-the-loop, stakeholders were often involved in a post-hoc evaluation manner instead of throughout the development process, and thus, have limited knowledge regarding the operating principle and potential weakness of the systems. Consequently, none of the existing LLMs-based innovations can be considered as being at Tier 3, which describes an AI system that is considered transparent for educational stakeholders (e.g., students, teachers, and parents).

\subsubsection{Privacy} 

The privacy issues related to LLMs-based innovations were rarely attended to or investigated in the reviewed studies. Specifically, for studies that have fine-tuned LLMs with textual data collected from students, none of these studies has explicitly explained their consenting strategies (e.g., whether students acknowledge the collection and intended usage of their data) and data protection measures (e.g., data anonymisation and sanitisation). This lack of attention to privacy issues is particularly concerning as LLMs-based innovations work with stakeholders' natural languages that may contain personal and sensitive information regarding their private lives and identities \citep{brown2022does}. It is possible that stakeholders might not be aware of their textual data (e.g., forum posts or conversations) on digital platforms (e.g., MOOCs and LMS) being used in LLMs-based innovations for different purposes of automation (e.g., automated reply and training chatbots) as the consenting process is often embedded into the enrollment or signing up of these platforms \citep{tsai2017learning}. This process can hardly be considered informed consent. Consequently, if stakeholders shared their personal information on these platforms in natural language (e.g., sharing phone numbers and addresses with group members via digital forums), such information could be used as training data for fine-tuning LLMs. This usage could potentially expose private information as LLMs are incapable of understanding the context and sensitivity of text, and thus, could return stakeholders' personal information based on semantic relationships \citep{brown2022does}.

%We only identified one study explicitly mentioning the privacy issues related to LLMs-based innovations \citep{merine2022risks}. In this study, the authors mentioned the need to consider privacy issues as such issues could expose students and teachers to cybercrime, where their personal information can be misused for malicious purposes (e.g., identity theft). Such privacy issues are particularly concerning as more than a half of the LLMs-based innovations (n=64) involved some form of data labelling during the development process, which increases the exposure of potentially personal and sensitive data (e.g., demographics and dialogues). In the Discussion section, we further elaborated on the lack of privacy considerations in the literature.

\subsubsection{Equality} 

Although most of the studies (n=95) used LLMs that only apply to English content, we also identified application scenarios of LLMs in automating educational tasks in 12 other languages. Specifically, 19 studies used LLMs that can be applied to Chinese content. Ten prior studies used LLMs for Vietnamese (n=3), Spanish (n=3), Italian (n=2), and German (n=2) contents. Additionally, seven studies applied LLMs to Croatian, Indonesian, Japanese, Romanian, Russian, Swedish, and Hindi content. While the dominance of English-based innovations remains a concerning equality issue, the availability of innovations that support a variety of other languages, specifically in non-western, educated, industrialized, rich and democratic (WEIRD) societies (e.g., Indonesia and Vietnam), may indicate a promising sign for LLMs-based innovations to have potential global impacts and levels such equality issues in the future. However, the financial burdens from adopting the state-of-the-art models (e.g., OpenAI's GPT-3 and Codex) could potentially exacerbate the equality issues, making the best-performing innovations only accessible and affordable to WEIRD societies.

\subsubsection{Beneficence}

A total of seven studies have discussed potential issues related to the violation of the ethical principle of beneficence. For example, one study has discussed the potential risk of adopting underperforming models, which could negatively affect students' learning experiences \citep{li2021natural}. Such issues could be minimised by deferring decisions made by such models \citep{schneider2022towards} and labelling the AI-generated content with a warning message (e.g., teachers' manual revision is mandatory before determining the actual correctness) \citep{angelone2022improved}. Apart from issues with adopting inaccurate models, two studies have suggested that potential bias and discrimination issues may occur if adopting a model that is accurate but unfair \citep{sha2021assessing,merine2022risks}. This issue is particularly concerning as most existing studies focused solely on developing an accurate model. Only nine reviewed studies released information regarding the descriptive data of different sample groups, such as gender and ethicality (e.g., \citep{pugh2021say}). Two studies have proposed potential approaches that could address such fairness issues. Specifically, using sampling strategies, such as balancing demographic distribution, has been found as an effective approach to improve both model fairness and accuracy \citep{sha2022leveraging,sha2022bigger}. These approaches are essential for ensuring that LLMs-based innovations will not perpetuate problematic and systematic biases (e.g., gender biases), especially as the best-performing LLMs are often black-boxed with little interpretability, traceability, and justification of the results \citep{wu2022analysis}.

\section{Discussion}

\subsection{Main Findings}

The current study systematically reviewed 118 peer-reviewed empirical studies that used LLMs to automate educational tasks. For the first research question (RQ1), we illustrated the current state of educational research on LLMs. Specifically, we identified 53 types of application scenarios of LLMs in automating educational tasks, summarised into nine general categories, including profiling and labelling, detection, assessment and grading, teaching support, prediction, knowledge representation, feedback, content generation, and recommendation. While some of these categories resonate with the utilities proposed in prior positioning works (e.g., feedback, content generation, and recommendation) \citep{kasneci2023chatgpt,rudolph2023chatgpt}, novel directions such as using LLMs to automate the creation of knowledge graph and entity further indicated the potential of LLMs-based innovations in supporting institutional practices (e.g., creating knowledge-based search engines across multiple disciplines). These identified directions could benefit from the state-of-the-art LLMs (e.g., GPT-3 and Codex) as most of the reviewed studies (92\%) focused on using BERT-based models, which often required manual effort for fine-tuning. Whereas, the state-of-the-art LLMs could potentially achieve similar performance with a zero-shot approach \citep{bang2023multitask}. While the majority of the reviewed studies (63\%) focused on using LLMs to automate classification tasks, there could be more future studies that aimed to tackle the automation of prediction and generation tasks with the more capable LLMs \citep{sallam2023utility}. Likewise, although supporting teachers are the primary focus (72\%) of the existing LLMs-based innovations, students and institutions could also benefit from such innovations as novel utilities could continue to emerge from the educational technology literature. Together, the findings of the first research question could spark educational researchers with ideas of exploring the potential of state-of-the-art LLMs in augmenting educational practices, specifically, the identified 53 types of application scenarios may all worth to re-explore in the light of ChatGPT and other powerful generative AI models \citep{kasneci2023chatgpt}.

Regarding the second research question (RQ2), we identified several practical challenges that need to be addressed for LLMs-based innovations to have actual educational benefits. The development and educational research on LLMs-based innovations are still in the early stages. Most of the innovations demonstrated a low level of technology readiness, where the innovations have yet to be fully integrated and validated in authentic educational contexts. This finding resonates with previous systematic reviews on related educational technologies, such as reviews on automated question generation \citep{kurdi2020systematic}, feedback provision \citep{cavalcanti2021automatic}, essay scoring \citep{ramesh2022automated}, and chatbot systems \citep{wollny2021we}. There is a pressing need for in-the-wild studies that provide LLMs-based innovations directly to educational stakeholders for supporting actual educational tasks instead of testing on different datasets or in laboratory settings. Such authentic studies could also validate whether the existing innovations can achieve the reported high model performance in real-life scenarios, specifically in prediction and generation tasks, instead of being limited to prior datasets. This validation process is vital for preventing inadequate usage, such as adopting a subject-specific prediction model for unintended subjects. Researchers need to carefully examine the extent of generalisability of their innovations and inform the limitations to stakeholders \citep{gavsevic2016learning}. However, addressing such needs could be difficult considering the current literature's poor replicability, which increases the barriers for others to adopt LLMs-based innovations in authentic educational contexts or validate with different samples. Similar replication issues have also been identified in other areas of educational technology research \citep{yan2022scalability}.

For the third research question (RQ3), we identified several ethical challenges regarding LLMs-based innovations. In particular, most of the existing LLMs-based innovations (92\%) were only transparent to AI researchers and practitioners (Tier 1), with only nine studies that can be considered transparent to educational technology experts and enthusiasts (Tier 2). The primary reason behind this low transparency can be attributed to the lack of human-in-the-loop components in prior studies. This finding resonates with the call for explainable and human-centred AI, which stresses the vital role of stakeholders in developing meaningful and impactful educational technology \citep{khosravi2022explainable,yang2021human}. Involving stakeholders during the development and evaluation of LLMs-based innovations is essential for addressing both practical and ethical issues. For example, as the current findings revealed, LLMs-based innovations are subject to data privacy issues but were rarely mentioned or investigated in the literature \citep{merine2022risks}, which may be due to the little voice that stakeholders had in prior research. The several concerning issues around beneficence also demand the involvement of stakeholders as their perspectives are vital for shaping the future directions of LLMs-based innovations, such as how responsible decisions can be made with these AI systems \citep{schneider2022towards}. Likewise, the equality issue regarding the financial burdens that may occur when adopting innovations that leverage commercial LLMs (e.g., GPT-3 and Codex) can also be further studied with institutional stakeholders. 

\subsection{Implications}

The current findings have several implications for education research and practice with LLMs, which we have summarised into three recommendations that aim to support future studies to develop practical and ethical innovations that can have actual benefits to educational stakeholders. First, the wide range of application scenarios of LLMs-based innovations can further benefit from the improvements in the capability of LLMs. Updating existing innovations with state-of-the-art LLMs may further reduce the amount of manual effort required for fine-turning and achieve similar performances \citep{bang2023multitask}. Considering the 53 identified use cases of LLMs in education, there are multiple research trajectories that could foster the development of practical educational technologies. These avenues have the potential to address some of the pressing challenges that plague the global education system. Particularly, the use cases involving teaching support, assessment and grading, feedback, and content generation categories (Table \ref{tab:educational-tasks}) could act as catalysts for the development of educational technologies that could alleviate teachers' workload and mental stress by automating the laborious tasks associated with creating, evaluating, and providing feedback for student assessments \citep{carroll2022teacher}. Similarly, further exploration of the use cases in profiling and labelling, detection, prediction, and recommendation could lead to the development of educational technologies that can deliver personalised learning support for each student across various disciplines \citep{wollny2021we}. Such improvements could enhance the overall well-being of teachers and increase students' learning opportunities, thereby contributing to the achievement of SDG 4 by 2030 \citep{boeren2019understanding}. Nonetheless, researchers should also be mindful of the potential financial and resource burdens that could be imposed on educational stakeholders when innovating with the commercial LLMs (e.g., GPT-3/4 and ChatGPT).

The unrivalled natural language generation capabilities exhibited by ChatGPT and other cutting-edge LLMs (e.g., LLaMA and PaLM 2) might also inspire future studies to delve into a broader spectrum of research directions. These include comparisons between the quality of student-generated and ChatGPT-generated writings \citep{li2023can} and evaluating these LLMs' capability to tackle educational assessments \citep{gilson2023does}. Such explorations would not only unveil the potential of LLMs and generative AI models in educational content generation and evaluation tasks but also expose the possible threats that these models pose to academic integrity, a pervasive issue across the education sector \citep{kasneci2023chatgpt}. Intriguingly, leveraging the use cases of LLMs in tasks such as creating knowledge representation \citep{zheng2023automatic} and classifying cognitive levels \citep{liu2022automated} could potentially facilitate the transition from outcome-focused to process-focused assessments. Here, LLMs and generative AI models could be employed for learning assessments in a manner similar to learning analytics \citep{gasevic2022towards}. Consequently, future studies may begin to explore methods of addressing the potential threats of LLMs with LLMs-based solutions.

For LLMs-based innovations to achieve a high level of technology readiness and performance, the current reporting standards must be improved. Future studies should support the initiative of open-sourcing their models/systems when possible and provide sufficient details about the test datasets, which are essential for others to replicate and validate existing innovations across different contexts, preventing the potential pitfall of another replication crisis \citep{maxwell2015psychology}. This initiative is particularly vital in the era of generative AI models as most of these models, especially the commercial ones (e.g., ChatGPT and the GPT series), are proprietary. Thus, when using these LLMs for augmenting educational practices, such as scoring student essays \citep{doewes2021limitations}, providing real-time feedback \citep{zheng2022effects}, or generating questions for learning activities \citep{sarsa2022automatic}, researchers need to be systematic and transparent about the reporting of the model usage and prompts \citep{wu2022analysis}. For example, when using the ChatGPT API for question generation at scale, researchers should at least report the exact models, prompts, and model temperature used in the process, as different models may differ in their ability to generate accurate and reliable content and the prompts are essential for others to replicate the same or similar results \citep{kasneci2023chatgpt}.

Apart from the aforementioned technical and methodological details, researchers and educational policymakers should also consider the potential wider impacts of LLMs-based solutions on different stakeholders. For example, in terms of detection and academic integrity, some institutions have rapidly adopted AI-detection tools that claim to have high accuracy and a low false positive rate. Yet, as disclosed in a recent report by Turnitin, a company whose AI-detection function has been utilised on more than 38.5 million student submissions, the real-world performance of their solution resulted in a significantly higher occurrence of false positives compared to their laboratory findings \citep{turnitin2023ai}. Such negligence can be devastating for students who have been falsely accused of academic misconduct, as well as for educators who must handle the repercussions. This example reinforced the importance of conducting rigorous scientific studies with key stakeholders when adopting any LLMs-based solutions that have direct or indirect impacts on students, educators, and other stakeholders. Likewise, the reporting of such studies should also adhere to high standards, incorporating both methodological specifics and detailed data descriptions. These details are especially pertinent when considering the diverse cultural backgrounds of students and the fact that most LLMs are primarily trained on English datasets, which could potentially introduce biases towards non-native English students \citep{liang2023gpt}.

Adopting a human-centred approach when developing and evaluating LLMs-based innovations are essential for ensuring these innovations remain ethical in practice, especially as ethical principles may not guarantee ethical AI due to their top-down manners (e.g., developed by regulatory bodies) \citep{mittelstadt2019principles}. Future studies need to consider the ethical issues that may arise from their specific application scenarios and actively involve stakeholders to identify and address such issues. Specifically, LLM-based innovations should aim to reach at least Tier 3 in the transparency index and TRL-7 in technology readiness. This involves a fully functional system being integrated into authentic learning environments and validated by students and educators in terms of its practicality and ethical considerations. For any decisions made by the LLM-based innovations, the relevant stakeholders should be informed about how the decision was reached, as well as the potential risks and biases involved. For instance, when students receive an assessment that has been automatically graded, these grades should be accompanied by a warning message indicating that they have been graded by LLMs and AI \citep{angelone2022improved}. Students should also have the opportunity to consult their teacher regarding any concerns.

The active involvement of stakeholders should also extend beyond the education sector, also involving policymakers and industry companies to establish the guidelines for adopting LLMs-based innovations in learning and teaching practices, as such adoptions could have broader implications on society beyond the education sector. For example, human-AI collaboration might become an essential skill for students to succeed in the job market as AI solutions become an integral component of productivity in the industrial sector \citep{wang2020human}. Therefore, institutions that aim to prohibit AI tools could inadvertently place their students at a disadvantage compared to other institutions that proactively welcome such changes. This could be achieved by consistently refining their policy regarding the use of LLMs and generative AI solutions, based on stakeholder feedback and empirical evidence.

\subsection{Limitations}

The current findings should be interpreted with several limitations in mind. First, although we assessed the practicality and ethicality of LLMs-based innovations with seven different items, there could be other aspects of these multi-dimensional concepts that we omitted. Nevertheless, these assessment items were chosen directly from the corresponding definitions and related to the pressing issues in the literature \citep{adams2021artificial,weidinger2021ethical}. Second, we only included English publications, which could have biased our findings regarding the availability of LLMs-based innovations among different countries. Thirdly, as we strictly followed the PRISMA protocol and only included peer-reviewed publications, we may have omitted the emerging works published in different open-sourced archives. These studies may contain interesting findings regarding the latest LLMs (e.g., ChatGPT). Additionally, this review focused on the potential of LLMs-based innovations in automating educational tasks, and thus, other pressing issues, such as the potential threat to academic integrity, were outside of the scope of this systematic scoping review. We briefly touched on these pressing issues in the implications and illustrated the importance of the current findings in supporting future educational studies to address these issues. Moreover, since this study is a systematic scoping review, we did not assess the quality of the included studies, and thus, the findings, particularly, the performance metrics extracted from the reviewed studies, may need further evaluation. The goal of this study is to provide an overview of the different educational tasks that can be augmented by LLMs and generative AI models, which can serve as a reference point for future studies to further develop on using the state-of-the-art models (e.g., ChatGPT and PaLM 2). Furthermore, the transparency index that we adopted for RQ3 did not consider the transparency to students, which could be an important direction for future human-centred AI studies. Finally, we recognise the rapid development in the field of artificial intelligence in education. It is pertinent to mention that a number of recent workshops and preliminary papers, while contributing to this field, were not incorporated in this scoping review due to time constraints \citep{leiker2023prototyping,ma2023better,caines2023application}. Their exclusion represents a limitation to the breadth of this study, acknowledging the relentless pace of scholarly advancements in this area.

\section{Conclusion}

In this study, we systematically reviewed the current state of educational research on LLMs and identified several practical and ethical challenges that need to be addressed in order for LLMs-based innovations to become beneficial and impactful. Based on the findings, we proposed three recommendations for future studies, including updating existing innovations with state-of-the-art models, embracing the initiative of open-sourcing models/systems, and adopting a human-centred approach throughout the developmental process. These recommendations could potentially support future studies to develop practical and ethical innovations that can be implemented in authentic contexts to automate a wide range of educational tasks.

\printendnotes

\bibliography{reference.bib}

\end{document}